\documentclass[runningheads]{llncs}
\usepackage[T1]{fontenc}
\usepackage{graphicx,verbatim}
\usepackage{amsmath}
\usepackage{amssymb} 
\begin{document}
\title{NeuroXVocal: Detection and Explanation of Alzheimer's Disease through Non-invasive Analysis of Picture-prompted Speech}
\titlerunning{NeuroXVocal}

\author{Nikolaos Ntampakis\inst{1,2}\orcidID{0009-0001-7546-4104} \and
Konstantinos Diamantaras\inst{1}\orcidID{0000-0003-1373-4022} \and
Ioanna Chouvarda\inst{3}\orcidID{0000-0001-8915-6658} \and
Magda Tsolaki\inst{4}\orcidID{0000-0002-2072-8010} \and
Vasileios Argyriou\inst{5}\orcidID{0000-0003-4679-8049} \and
Panagiotis Sarigianndis\inst{6,2}\orcidID{0000-0001-6042-0355}}

\authorrunning{N. Ntampakis et al.}

\institute{International Hellenic University, Sindos, Greece \and
MetaMind Innovations, Kozani, Greece \and
Aristotle University of Thessaloniki, Thessaloniki, Greece \and
Greek Association of Alzheimer’s Disease \& Related Disorders, Thessaloniki, Greece \and
Kingston University London, London, UK \and
University of Western Macedonia, Kozani, Greece
}

\maketitle
\begin{abstract}
The early diagnosis of Alzheimer's Disease (AD) through non invasive methods remains a significant healthcare challenge. We present NeuroXVocal, a novel dual-component system that not only classifies but also explains potential AD cases through speech analysis. The classification component (Neuro) processes three distinct data streams: acoustic features capturing speech patterns and voice characteristics, textual features extracted from speech transcriptions, and precomputed embeddings representing linguistic patterns. These streams are fused through a custom transformer-based architecture that enables robust cross-modal interactions. The explainability component (XVocal) implements a Retrieval-Augmented Generation (RAG) approach, leveraging Large Language Models combined with a domain-specific knowledge base of AD research literature. This architecture enables XVocal to retrieve relevant clinical studies and research findings to generate evidence-based context-sensitive explanations of the acoustic and linguistic markers identified in patient speech. Using the IS2021 ADReSSo Challenge benchmark dataset, our system achieved state-of-the-art performance with 95.77\% accuracy in AD classification, significantly outperforming previous approaches. The explainability component was qualitatively evaluated using a structured questionnaire completed by medical professionals, validating its clinical relevance. NeuroXVocal's unique combination of high-accuracy classification and interpretable, literature-grounded explanations demonstrates its potential as a practical tool for supporting clinical AD diagnosis.

\keywords{Alzheimer  \and Multimodal \and Explainable Healthcare AI.}
\end{abstract}

\section{Intoduction}
Alzheimer's Disease (AD) has emerged as a critical global health concern, affecting over 55 million people worldwide with nearly 10 million new cases annually~\cite{who}. Early detection through non-invasive methods remains crucial for effective intervention and treatment planning. While traditional diagnostic approaches rely on neuroimaging or invasive procedures, recent advances in artificial intelligence have opened new possibilities for early detection through speech analysis~\cite{wong,luz}. This paper presents NeuroXVocal, a novel dual-component system that not only classifies but also explains its diagnostic predictions through speech analysis of patients describing images, whether they are identified as having Alzheimer's disease or being cognitively healthy.
The relationship between cognitive decline and speech patterns has been extensively studied using the ADReSSo benchmark dataset ~\cite{adr}. Syed et al. achieved significant results using functionals of deep textual embeddings, reporting 84.51\% accuracy in AD detection~\cite{syed}. Shah et al. further investigated language-agnostic speech representations, demonstrating the effectiveness of speech intelligibility features with 79.6\% accuracy~\cite{shah}. More recently, Fu et al. proposed a multimodal fusion method combining acoustic and semantic information using ImageBind audio encoder and ELMo, achieving 90.3\% accuracy~\cite{fu}. Li et al. demonstrated promising results using Whisper-based transfer learning, achieving 84.51\% accuracy and 84.50\% F1-score by innovatively using full transcripts as prompts during fine-tuning~\cite{li}. The latest advancement by Lee et al. introduced a graph neural network leveraging image-text similarity from vision language models, achieving 88.73\% accuracy~\cite{lee}.
While these approaches have shown promising results in classification, the field has seen limited progress in explaining the reasoning behind diagnostic predictions. Recent work by Iqbal et al. employed Local Interpretable Model-agnostic Explanations (LIME) and SHapley Additive exPlanations (SHAP) to provide insights into linguistic markers of cognitive decline~\cite{iqbal}. Similarly, Bang et al. explored the use of LLMs for generating evidence-based explanations of speech patterns, though their approach was limited by the interpretability of the underlying language model~\cite{bang}. However, these studies still face challenges in providing comprehensive, clinically-actionable explanations that bridge the gap between machine learning predictions and medical decision-making.
Building upon these foundations, we present NeuroXVocal, a novel framework that advances the state-of-the-art in both AD detection accuracy and clinical interpretability. Our main contributions can be summarized as follows:
\begin{enumerate}
    \item Development of an end-to-end holistic framework that seamlessly integrates classification and explanation for diagnosis of potential AD patients.
    \item Introduction of a novel explainability component (XVocal), bridging the gap between machine learning predictions and clinical interpretability. Qualitative evaluation through structured questionnaires completed by medical professionals validated XVocal's ability to produce clear and clinically relevant explanations, validating its potential as a reliable clinical decision support tool.
    \item Achievement of state-of-the-art performance with 95.77\% accuracy in classification (Neuro) on the IS2021 ADReSSo Challenge benchmark dataset, significantly outperforming existing approaches
\end{enumerate}

\section{Methodology}
\begin{figure}
\centering
\includegraphics[width=12cm,height=12cm]{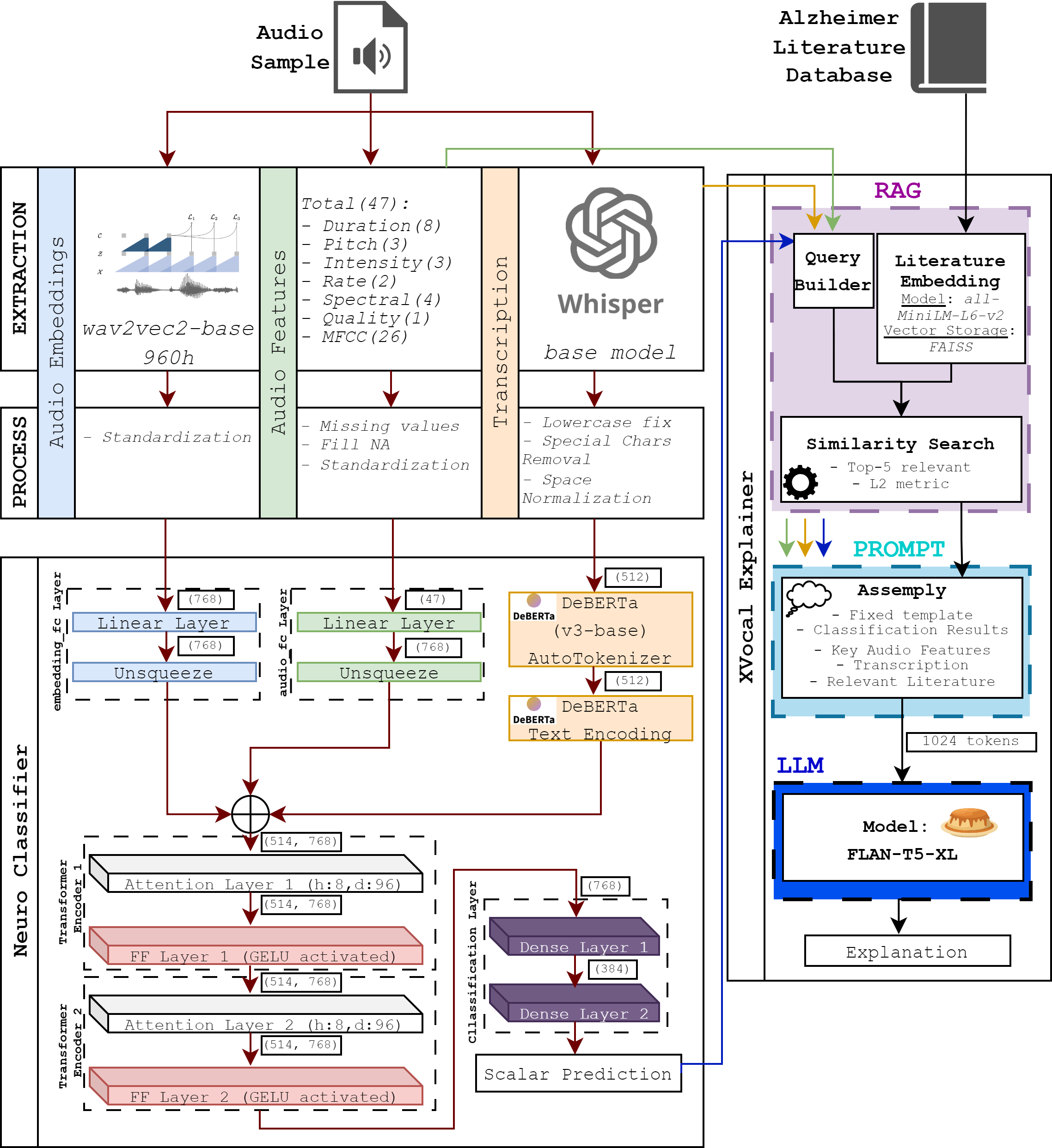}
\caption{NeuroXVocal Architecture} \label{arch}
\end{figure}

Our proposed novel NeuroXVocal system consists of two primary components, as in Fig.~\ref{arch}: (1) the Neuro classifier for AD detection through multimodal analysis of speech data, and (2) the XVocal explainer for generating clinically-interpretable justifications. The system processes input audio samples through multiple parallel streams to extract complementary features before fusion and classification.
\subsection{Feature Extraction and Processing}
Let $x$ be an input audio sample. From this input, we extract three distinct feature representations. The acoustic features $f_a(x) = \phi_a(x) \in \mathbb{R}^{47}$ comprise temporal characteristics (speech/pause ratios), prosodic features (pitch, intensity), articulation metrics, spectral properties, voice quality indicators (jitter, shimmer, harmonics-to-noise ratio), and 13 Mel Frequency Cepstral Coefficients(MFCC) coefficients with their standard deviations. These features (47 int total) undergo standardization and missing value imputation.
For speech embeddings, we employ Wav2Vec2-base-960h~\cite{baevski} after converting audio to mono and resampling to 16kHz:
\begin{equation}
f_e(x) = \text{Mean}(\text{Wav2Vec2}(\text{Preprocess}(x))) \in \mathbb{R}^{768}
\end{equation}
where the embeddings are standardized before further processing.
The textual features are obtained using Whisper ASR~\cite{radford} for transcription followed by DeBERTa-v3-base~\cite{he} encoding:
\begin{equation}
f_t(x) = \text{DeBERTa}(\text{Preprocess}(\text{Whisper}(x))) \in \mathbb{R}^{768}
\end{equation}
where preprocess includes lowercase conversion, special character removal, and space normalization.
\subsection{Neuro Classifier}
The classification component implements a novel fusion architecture. We first project the acoustic and speech embedding features to a common dimensional space:
\begin{equation}
h_a = \text{Linear}(f_a(x)) \in \mathbb{R}^{768}, \quad h_e = \text{Linear}(f_e(x)) \in \mathbb{R}^{768}
\end{equation}
The fusion process concatenates these projections with the text embeddings in the projected dimentiontins of $514\times768$:
\begin{equation}
H = [h_a; h_e; f_t(x)] \in \mathbb{R}^{514 \times 768}
\end{equation}
A two-layer transformer encoder processes this representation. Each layer implements multi-head attention with 8 heads, where the input $H$ is projected to queries ($Q$), keys ($K$), and values ($V$):
\begin{equation}
\text{Attention}(Q,K,V) = \text{Concat}(\text{head}_1,...,\text{head}_8)W^O
\end{equation}
where $W^O$ is the output projection matrix, and each attention head is computed as:
\begin{equation}
\text{head}_i = \text{softmax}(\frac{QW^Q_i(KW^K_i)^T}{\sqrt{d_k}})VW^V_i
\end{equation}
where $W^Q_i, W^K_i, W^V_i \in \mathbb{R}^{768 \times 96}$ are learned parameter matrices, and $d_k = 96$ is the head dimension.This is followed by a feed-forward network with GELU activation:
\begin{equation}
Z = \text{FFN}(\text{Attention}(H)) \in \mathbb{R}^{514 \times 768}
\end{equation}
The final classification uses a two-layer classifier:
\begin{equation}
p(y|x) = \sigma(\text{Dense}(z_0))
\end{equation}
where $\sigma$ is the sigmoid activation function for binary classification.
\subsection{XVocal Explainer}
The novel explainability component implements a RAG approach that processes the extracted features along with the Neuro classifier's prediction. The explanation generation begins by constructing a structured prompt query $q$ through a template:
\begin{equation}
q = \{\text{class}(p(y|x)) \oplus \text{features}(f_a(x)) \oplus \text{speech}(f_e(x)) \oplus \text{transcript}(f_t(x))\}
\end{equation}
The relevant literature corpus $\mathcal{L}$ is preprocessed into semantic chunks by splitting each document into paragraphs and then into individual sentences to create a fine-grained context pool $\{c_1, ..., c_n\}$. Using all-MiniLM-L6-v2~\cite{wang}, we construct a dense vector index:
\begin{equation}
\mathbf{E}_c = \{\text{MiniLM}(c_i) \in \mathbb{R}^{384} | c_i \in \mathcal{L}\}
\end{equation}
where each chunk is encoded into a 384-dimensional embedding space. These embeddings are indexed using FAISS~\cite{douze} L2 distance metric:
\begin{equation}
\mathcal{I} = \text{FAISS}_{\text{L2}}(\mathbf{E}_c)
\end{equation}
For retrieval, the query $q$ is encoded in the same embedding space and the top-5 most relevant chunks are retrieved using nearest neighbor search:
\begin{equation}
\mathcal{L}_r = \{\mathcal{I}.\text{search}(\text{MiniLM}(q), k=5)\}
\end{equation}
The final explanation is generated using FLAN-T5-XL~\cite{chung}:
\begin{equation}
E = \text{FLAN-T5}(q \oplus \mathcal{L}_r; \tau, p)
\end{equation}
where $\tau$ and $p$ are the temperature and top-p sampling parameters respectively, controlling the generation coherence.

\section{Experiments and Results}

\subsection{Implementation Details}

All experiments were conducted on Ubuntu using 8xNVIDIA A16 GPUs with 126GB system RAM. The Neuro classifier trained for a maximum of 200 epochs. For the XVocal component, we used FAISS (CPU) for retrieval and deployed using 4-bit quantization.  Each training round was completed on an average of 9 hours in our setup.

\subsection{Dataset}

We utilised the ADReSSo Challenge dataset~\cite{adr} for the probable AD prediction task. The data is organized in the diagnosis folder, with 166 patients in the training set (79 cognitively normal [cn], 87 probable Alzheimer's disease [ad]) and 71 patients in the test set. The test set is kept independent for transparent evaluation. The dataset is accessible through DementiaBank membership, requiring registration and administrator approval.

\subsection{Results}
\begin{table}
\centering
\caption{Performance comparison between methodologies.}\label{tab1}
\begin{tabular}{|l|c|l|c|}
\hline
Methodology &  5-fold Accuracy($\pm$ std\%) & Acc[\%] & F1-score[\%]\\
\hline
Syed et al.(2021)~\cite{syed} &  & 84.51\% & 84.45\%\\
Shah et al.(2023)~\cite{shah} &  & 79.60\% & \\
Fu et al.(2024)~\cite{fu} &  & 90.3\% & 91.4\%\\
Li et al.(2024)~\cite{li} &  & 84.51\% & 84.5\%\\
Lee et al.(2025)~\cite{lee} &  & 88.73\% & 88.23\%\\
\textbf{NeuroXVocal (Neuro Classifier)} & \textbf{96.24\% $\pm$ 2.47\%} & \textbf{95.77\%} & \textbf{95.76\%}\\
\hline
\end{tabular}
\end{table}
Regarding the results of the Neuro Classifier incorporated in our NeuroXVocal methodology, we compared with prominent and recent state-of-the-art methodologies as shown in Table~\ref{tab1}. To evaluate the performance, we utilized the widely adopted accuracy and F1-score metrics, defined as:
\begin{align}
    \text{Accuracy} = \frac{TP + TN}{TP + TN + FP + FN} \qquad & \text{F1-score} = \frac{2TP}{2TP + FP + FN}
\end{align}
where TP (True Positives) represents correctly identified AD cases, TN (True Negatives) represents correctly identified control cases, FP (False Positives) represents control cases incorrectly classified as AD, and FN (False Negatives) represents AD cases incorrectly classified as control. As demonstrated in Table~\ref{tab1}, our Neuro classifier achieved robust performance across multiple evaluation scenarios. In the 5-fold cross-validation setting, we obtained an average accuracy of 96.24\% with a standard deviation of 2.47\%. When trained on the full training set and evaluated on the independent test set, our method achieved 95.77\% accuracy and 95.76\% F1-score, substantially outperforming all previous approaches.
\begin{table}
\centering
\caption{Criteria and expert evaluation results for XVocal's explanations}\label{tab2}
\begin{tabular}{|p{4cm}|p{6cm}|c|}
\hline
Assessment Focus & Scale & Mean Score \\
\hline
Clear justification of diagnosis & 1-Not clear, 5-Very clear & 3.96 \\
Pertinence of identified markers & 1-Not relevant, 5-Highly relevant & 3.85 \\
Consistency with medical knowledge & 1-No alignment, 5-High alignment & 3.63 \\
Explanation-based confidence & 1-Not confident, 5-Highly confident & 3.63 \\
Recognition of disease indicators & 1-No markers identified, 5-Highly appropriate markers identified & 3.98 \\
Utility for diagnosis & 1-Not useful, 5-Highly useful & 3.70 \\
Coherence and plausibility & 1-Not sound, 5-Very sound & 3.74 \\
Expected consensus & 1-Very unlikely, 5-Very likely & 3.56 \\
Robustness of reasoning & 1-Not at all plausible, 5-Highly plausible & 3.77 \\
Potential for misinterpretation & 1-Not misleading, 5-Highly misleading & 2.38 \\
\hline
\end{tabular}
\end{table}

To assess the clinical relevance and utility of XVocal's explanations, we conducted a comprehensive qualitative evaluation with medical experts. Each expert evaluated explanations for 20 patient cases (10 AD, 10 CN) using a structured questionnaire with 10 criteria, rated on a 5-point Likert scale.  For the knowledge base of the RAG component, we have incorporated a curated corpus of 10 seminal publications~\cite{fraser,toth,ahmed,forbes,taler,konig,ding,qi,chen,agbavor} covering linguistic markers, spontaneous speech analysis, and LLM applications in AD detection.

The evaluation results (Table \ref{tab2}) demonstrate strong performance across multiple dimensions of clinical utility. XVocal achieved notably high scores in AD marker identification (3.98) and explanation clarity (3.96), indicating its effectiveness in highlighting relevant diagnostic features. The system also performed well in identifying relevant linguistic features (3.85) and maintaining logical soundness (3.74), suggesting reliable diagnostic reasoning.
Particularly noteworthy is the low score for potentially misleading aspects (2.38), indicating that experts found minimal risk of misinterpretation in XVocal's explanations. This is crucial for clinical applications where accuracy and reliability are paramount. The system also demonstrated good alignment with clinical understanding (3.63) and strong utility for supporting diagnostic decisions (3.70).

\section{Ablation Study}

To assess the contribution of each modality, we conducted systematic experiments by removing components and adapting the network architecture accordingly. For each combination, we modified the dimensions of the input layers to match the sizes of the feature vector. The fusion layer and attention mechanisms were adjusted proportionally while maintaining the core architecture design.

\begin{table}
\centering
\caption{Ablation study results showing modality combinations}\label{tab3}
\begin{tabular}{|c|c|c|l|l|}
\hline
Audio & Audio & Text & & \\
Embed. & Feat. & Trans. & Accuracy[\%] & F1-score[\%] \\
\hline
\checkmark & \checkmark & \checkmark & 95.77 & 95.76 \\
 & \checkmark & \checkmark & 89.86 & 89.86 \\
\checkmark & & \checkmark & 91.30 & 91.29 \\
\checkmark & \checkmark &  & 84.78 & 84.70 \\
\hline
\end{tabular}
\end{table}

Results, as shown in Table \ref{tab3}, demonstrate the synergistic effect of multimodal fusion, with transcription features providing the strongest individual contribution when combined with audio embeddings (91.30\%). The transcription features prove crucial, as configurations lacking this component show reduced performance (84.78\%). Acoustic features seems to be the weaker modality, suggesting they capture complementary speech characteristics. The optimal performance (95.77\%) achieved with all three modalities indicates each component contributes unique discriminative information essential for robust AD detection.

\section{Conclusion}

We presented NeuroXVocal, a novel dual-component system for AD detection and explanation that achieved state-of-the-art performance (95.77\% accuracy, 95.76\% F1-score) on the ADReSSo Challenge benchmark, significantly outperforming existing approaches. The multimodal architecture of the system demonstrated robust capability for extraction and fusion of features, with ablation studies confirming the complementary nature of the modalities. The RAG-based explainability component, validated through expert evaluation, successfully bridges the gap between ML predictions and clinical interpretability.
Future work will focus on developing a real-time inference pipeline and implementing streaming audio processing for immediate feature extraction. We plan to extend the system with an interface for clinical deployment, incorporating incremental learning capabilities to adapt to new data patterns. Additionally, we aim to expand the knowledge base with continuous literature updates and enhance the RAG component with domain-specific prompt engineering for more targeted explanations. Further validation through large-scale clinical trials will help establish NeuroXVocal's efficacy as a practical diagnostic support tool.

%
%
%

\begin{thebibliography}{8}
\bibitem{who}
World Health Organization.: Global status report on the public health response to dementia. WHO Press, Geneva (2021)
\bibitem{wong}
Wong, K.K., et al.: Enhance early diagnosis accuracy of Alzheimer's disease by elucidating interactions between amyloid cascade and tau propagation. In: MICCAI 2023, pp. 1–10. Springer, Vancouver (2023).
\bibitem{luz}
Luz, S., Haider, F., de la Fuente, S., et al.: Noninvasive automatic detection of Alzheimer's disease through speech analysis. In: Frontiers in Aging Neuroscience, vol. 15, Article 1224723 (2023)
\bibitem{adr}
Luz, S., Haider, F., de la Fuente, S., Fromm, D., MacWhinney, B. (2021).: ADReSSo Challenge Dataset [Data set]. DementiaBank. Retrieved January 9, 2025, from https://dementia.talkbank.org/ADReSSo-2021/
\bibitem{syed}
Syed, Z.S., Shah, M.S., Lech, M., et al.: Tackling the ADRESSO Challenge 2021: The MUET-RMIT System for Alzheimer's Dementia Recognition from Spontaneous Speech. In: Interspeech 2021, pp. 3780--3784. ISCA, Brno (2021)
\bibitem{shah}
Shah, Z., Sawalha, J., Tasnim, M., et al.: Exploring language-agnostic speech representations using domain knowledge for detecting Alzheimer's dementia. In: ICASSP 2023, pp. 1--5. IEEE Press, Rhodes (2023)
\bibitem{fu}
Fu, Y., Xu, L., Zhang, Y., et al.: Classification and diagnosis model for Alzheimer's disease based on multimodal data fusion. Medicine \textbf{103}(52), e41016 (2024)
\bibitem{li}
Li, J., Zhang, W.Q.: Whisper-Based Transfer Learning for Alzheimer Disease Classification: Leveraging Speech Segments with Full Transcripts as Prompts. In: ICASSP 2024, pp. 11211--11215. IEEE Press, Seoul (2024)
\bibitem{lee}
Lee, B., Bang, J.U., Song, H.J., et al.: Alzheimer's disease recognition using graph neural network by leveraging image-text similarity from vision language model. Scientific Reports \textbf{15}, 997 (2025)
\bibitem{iqbal}
Iqbal, F., Syed, Z.S., Syed, M.S.S., Syed, A.S.: An Explainable AI Approach to Speech-Based Alzheimer's Detection Using Linguistic Features. In: ISCA SMM 2024, pp. 1--6. ISCA Press (2024)
\bibitem{bang}
Bang, J.-U., Han, S.-H., Kang, B.-O.: Alzheimer's Disease Recognition from Spontaneous Speech Using Large Language Models. In: ETRI Journal \textbf{46}(1), pp. 1--10 (2024)
\bibitem{baevski}
Baevski, A., Zhou, H., Mohamed, A., Auli, M.: wav2vec 2.0: A framework for self-supervised learning of speech representations. Advances in Neural Information Processing Systems \textbf{33}, 12449–12460 (2020)
\bibitem{radford}
Radford, A., Kim, J. W., Hallacy, C., et al.: Robust Speech Recognition via Large-Scale Weak Supervision. OpenAI (2022)
\bibitem{he}
He, P., Gao, J., Chen, W.: DeBERTaV3: Improving DeBERTa using ELECTRA-Style Pre-Training with Gradient-Disentangled Embedding Sharing. International Conference on Learning Representations (ICLR) (2023)
\bibitem{douze}
Douze, M., Guzhva, A., Deng, C., Johnson, J., Szilvasy, G., et al.: The Faiss Library. IEEE Transactions on Big Data \textbf{7}(3), 535–547 (2024)
\bibitem{wang}
Wang, W., Wei, F., Dong, L., Bao, H., Yang, N., Zhou, M.: MiniLM: Deep Self-Attention Distillation for Task-Agnostic Compression of Pre-Trained Transformers. Advances in Neural Information Processing Systems \textbf{33}, 5776–5788 (2020)
\bibitem{chung}
Chung, H. W., Hou, L., Longpre, S., et al.: Scaling Instruction-Finetuned Language Models. In: Proceedings of the Association for Computational Linguistics (ACL), pp. 1–12 (2023)
\bibitem{fraser}
Fraser, K.C., Meltzer, J.A., Rudzicz, F.: Linguistic features identify Alzheimer's disease in narrative speech. Journal of Alzheimer's Disease \textbf{49}(2), 407--422 (2016)
\bibitem{toth}
Tóth, L., et al.: Speech recognition-based classification of mild cognitive impairment and dementia. LNCS \textbf{11096}, 89--98. Springer, Heidelberg (2018)
\bibitem{ahmed}
Ahmed, S., Haigh, A.M., de Jager, C.A., Garrard, P.: Connected speech as a marker of disease progression in autopsy-proven Alzheimer's disease. Brain \textbf{136}(12), 3727--3737 (2013)
\bibitem{forbes}
Forbes-McKay, K.E., Shanks, M.F., Venneri, A.: Charting the decline in spontaneous writing in Alzheimer's disease: a longitudinal study. Acta Neuropsychiatrica \textbf{25}(6), 320--327 (2013)
\bibitem{taler}
Taler, V., Phillips, N.A.: Language performance in Alzheimer's disease and mild cognitive impairment: A comparative review. Journal of Clinical and Experimental Neuropsychology \textbf{30}(5), 501--556 (2008)
\bibitem{konig}
König, A., et al.: Automatic Speech Analysis for the Assessment of Alzheimer's Disease: A Review. Journal of Alzheimer's Disease Reports \textbf{2}(1), 1--15 (2018)
\bibitem{ding}
Ding, K., Chetty, M., Noori Hoshyar, A., et al.: Speech based detection of Alzheimer's disease: a survey of AI techniques, datasets and challenges. Artificial Intelligence Review \textbf{57}, 325 (2024)
\bibitem{qi}
Qi, X., Zhou, Q., Dong, J., Bao, W.: Noninvasive automatic detection of Alzheimer's disease from spontaneous speech: a review. Frontiers in Aging Neuroscience \textbf{15}, Article 1224723 (2023)
\bibitem{chen}
Chen, J., Ye, J., Tang, F., Zhou, J.: Automatic Detection of Alzheimer's Disease Using Spontaneous Speech Only. In: Interspeech 2021, pp. 3830--3834 (2021)
\bibitem{agbavor}
Agbavor, F., Liang, H.: Predicting dementia from spontaneous speech using large language models. PLOS Digital Health \textbf{1}(12), e0000168 (2022)
\end{thebibliography}
%

\end{document}